**Vision-Language Assistant for Emotional Reactions to Risky Driving**

**Harine Choi (ORCID: 0009-0009-1865-0312)**
Department of Computer Science and Engineering,
Texas A&M University, 77840, College Station, Texas, United States
Email: harinec@tamu.edu

**Eun Hak Lee (ORCID: 0000-0001-5989-4895)* Corresponding Author**
School of Railway Operation Systems,
Kyungil University, Gyeongsan-si, Gyeongsangbuk-do, Republic of Korea
Email: eunhak@kiu.ac.kr

**Zhengzhong Tu (ORCID: 0000-0002-7594-2292)**
Department of Computer Science and Engineering,
Texas A&M University, 77840, College Station, Texas, United States
E-mail: tzz@tamu.edu

**ABSTRACT**
This study introduces a vision–language pipeline that detects risky driving behaviors and generates emotionally expressive responses to support driver awareness and comfort. Although vision–language models have advanced perception and reasoning in autonomous driving, existing systems rarely consider the emotional dimension or real-world user experience. Keep Yelling Assistant (KYA) detects high-risk driving maneuvers in real time, such as sudden cut-ins. It then produces emotional responses through a large language model tailored to driver preferences. The framework comprises two core modules. The vision module uses YOLOv8 variants to detect nearby vehicles and identify risky behaviors such as sudden cut-ins. Key driving metrics, including relative distance, speed, and projected reach time, are extracted and normalized to produce a structured behavior log. The language module processes this log with user-defined emotional tone settings (e.g., neutral, humorous, analytical) and generates verbal reactions using state-of-the-art LLMs (ChatGPT-4o, Claude 3, Gemini 2.5, and Copilot). We evaluated the proposed system using dashcam videos containing risky driving behaviors and a user study involving 108 participants. Participants selected preferred response styles, and LLMs were evaluated based on emotional alignment. All models received favorable ratings, though preferences varied across personas. Notably, the combination of YOLOv8s and ChatGPT-4o achieved the highest score, 4.29 out of 5.00. By integrating real-world perception with emotionally adaptive dialogue, KYA introduces a new paradigm for emotionally intelligent in-vehicle AI. It offers promising directions for improving safety, trust, and emotional well-being in both conventional and autonomous vehicles.

**INTRODUCTION**

Modern urban driving environments are increasingly characterized by risky behaviors such as sudden lane changes without signaling, tailgating in congested traffic, or excessive speeding through intersections (1). These actions are not only statistically linked to a higher incidence of rear-end and side-swipe collisions, but also evoke intense emotional reactions from human drivers, ranging from irritation to panic (1). In the United States, the National Highway Traffic Safety Administration (NHTSA) attributes over 50% of fatal crashes to various forms of risky driving, including speeding and improper lane changes (*3*). In China, recent studies based on urban dashcam footage report that nearly 40% of high-risk interactions, particularly at intersections, are triggered by driver misbehavior rather than environmental conditions (*4*). In South Korea, over 30% of reported urban traffic accidents involve illegal maneuvers or dangerous driving actions, many of which fall under the category of risky driving as defined by contemporary traffic psychology frameworks (*5*).

These behaviors not only compromise traffic safety but also elicit emotionally charged responses from surrounding drivers, such as shouting, gesturing, or retaliatory driving. For example, Morgado et al. (*6*) emphasized that road safety campaigns incorporating emotional content can better resonate with drivers and reduce risky behavior. Chand and Karthikeyan (*7*) found that emotion-aware drowsiness detection systems can significantly reduce the risk of accidents by providing timely alerts. Liu et al. (*8*) showed that empathetic in-vehicle systems, which infer driver emotions from behavior and traffic context, can improve comfort and mitigate potentially dangerous reactions. Similarly, Li et al. (9) demonstrated that intelligent cockpits capable of auditory emotion regulation help manage extreme emotional states like anger and prevent safety-critical incidents. Collectively, these studies highlight the importance of recognizing and responding to driver emotions in real time. Emotional engagement, whether through system feedback or virtual assistants, can help reduce stress, prevent escalation, and enhance driver awareness. This ultimately contributes to safer road environments.

Recent years have seen growing interest in integrating emotion recognition and sentiment analysis into intelligent vehicle systems to enhance both safety and user experience (*9-18*). For example, Jadhav (*10*) proposed a multimodal framework combining speech, text, and visual cues to detect driver's emotions in real time. Pandya and Thakkar (*11*) showed that LSTM models effectively classify sentiments in self-driving car datasets. Huang et al. (*12*) demonstrated that emotionally aligned AI voice assistants can improve driver focus and mood. Wang et al. (*13*) introduced the ViE-Take dataset to study emotion during takeover events in autonomous driving. Stappen et al. (*14*) developed the MUSE-CAR dataset for sentiment analysis using in-vehicle speech and visuals, while Zou et al. (*15*) evaluated how audio-visual feedback affects emotional response. Wachter (*16*) proposed emotion tracking and reward mechanisms to encourage positive driving behavior. Finally, Giri et al. (*17*) and Zhang et al. (*18*) extended sentiment detection to the external driving scene, highlighting the role of affect recognition in both driver support and scene interpretation.

Vision-language models (VLMs) have recently gained attention in autonomous driving research for their potential to enhance scene understanding, decision-making, and natural language interaction (*19-28*). For example, SimpleLLM4AD (*19*) integrates visual prompts with graph-based reasoning for direct driving control, while GPT-Driver aligns LLM-generated motion plans with human trajectories. DriveDreamer-2 (*20*) and SurrealDriver (*21*) explore generative agents capable of simulating realistic driving behavior and internal cognitive processes. Another emerging direction is cooperative and human-in-the-loop driving, where models like CoLMDriver (*22*) employ language-based negotiation with surrounding agents, and ReasonDrive (*23*) or Reason2Drive (*24*) apply chain-of-thought reasoning for explainable decision-making. Meanwhile, visual question answering remains a core application, with systems like DriveLM, LingoQA, Nuscenes-QA, and NuPlanQA enabling interpretable understanding of traffic scenes through structured queries (*25-28*). To ensure robustness, recent studies have benchmark model performance against human baselines across diverse scenarios. Frameworks like Driving with Regulation incorporate rule-aware retrieval to align VLMs outputs with traffic laws and safety constraints. Collectively, these developments underscore the evolving role of VLMs as reasoning engines for transparent, interactive, and cognitively enriched autonomous driving. These models reflect a shift from rigid perception–planning pipelines toward more interpretable, language-driven agents. By enabling language-based interaction, VLMs offer human-aligned, transparent representations of vehicle behavior.

Despite these advances, most LLM-based advanced driver assistance systems (ADAS) remain limited to neutral or task-driven outputs (*30*). Specifically, emotional states and psychological responses, which are central to human driving experiences, remain largely unaddressed. Neglecting these core elements of how humans perceive and react to risk limits both the realism and the effectiveness of human-vehicle interaction. Emotionally neutral systems may correctly identify hazards yet fail to convey urgency or reassurance, leading to user disengagement or mistrust.

To address this gap and improve safety, comfort, and trust, this study presents an emotional reaction assistance system, the Keep Yelling Assistant (KYA), a vision–language framework designed to generate emotionally expressive responses to risky driving interactions. KYA interprets risky driving behaviors and produces persona-aligned reactions that reflect how human drivers might perceive and respond in hazardous traffic situations. By articulating emotional responses on behalf of the driver, the system aims to support affect-aware interaction during stressful moments. Importantly, KYA establishes an integrated perception–behavior–language architecture. The system consists of three core components. First, a YOLO-based perception module detects behavior-specific risk patterns, such as sudden cut-ins. Second, a behavioral analysis module interprets the interaction context based on structured driving metrics. Third, a large language model generates natural language responses that convey emotions.

The contributions of this work are as follows. 1) We propose a vision-language framework design for efficient and interpretable response generation. 2) We develop a practical vision module that detects cut-in behaviors from real dashcam footage using lightweight models. 3) We implement a language model that generates persona-aligned emotional responses based on user preference.

## METHODS

### Overview

This study introduces the KYA framework, which analyzes and reacts to risky driving behaviors in real-world dashcam videos, with empirical evaluation specifically focused on sudden cut-in scenarios. The system is designed to simulate emotional responses on behalf of the driver, offering both situational awareness and emotionally personalized feedback. The pipeline consists of two main components: 1) a vision model for detecting and analyzing driving behaviors, and 2) a language model for generating tailored emotional and linguistic responses. Together, these modules convert raw driving footage into meaningful, personality-aware feedback.

As illustrated in **Figure 1**, the pipeline begins with the vision module, which takes raw dashcam footage as input, including high-risk driving scenes curated from real-world platforms like YouTube. Using advanced computer vision techniques such as YOLOv8, the module performs several preprocessing tasks including frame sampling, resolution normalization, and scene segmentation. From this processed video, the system identifies nearby vehicles and classifies their characteristics (e.g., vehicle type, color, speed). In this study, the behavior recognition component was operationalized specifically for sudden cut-in events, identified through spatiotemporal proximity patterns. To quantify interaction intensity, the system calculates dynamic metrics including relative speed, inter-vehicle distance, and projected reach time (PRT). These indicators serve as relative behavioral risk cues and are recorded in a structured behavior log, which provides input to the language module.

The language module takes over by ingesting the structured behavior log along with the driver's personal preference settings, which define the desired emotional tone (e.g., neutral, humorous, angry, analytical). Using state-of-the-art LLMs such as ChatGPT-4o, Claude 3, Gemini 2.5, or Copilot, the model performs high-level reasoning tasks. These include interpreting the situation, assessing potential risk, and re-framing the incident from a human driver's perspective. Based on this analysis, the model generates an emotionally resonant narrative, aligned with the user’s chosen persona (e.g., neutral, angry, or even diplomatically forgiving). The output is rendered in natural language and passed to a Text-to-Speech (TTS) engine, which produces spoken feedback with a voice tone tailored to match the selected emotional style. The system supports diverse vocal reactions, ranging from dry neutral to supportive encouragement. All interactions and outputs are logged to support future refinement and model improvement.

This pipeline integrates a vision model and a language model to connect perceptual risk cues with emotionally contextualized response generation. Specifically, it establishes an integrated perception–behavior–language architecture. By structuring spatiotemporal driving cues into interpretable behavioral indicators and coupling them with large language model reasoning, the system bridges low-level perception and high-level affective interaction.

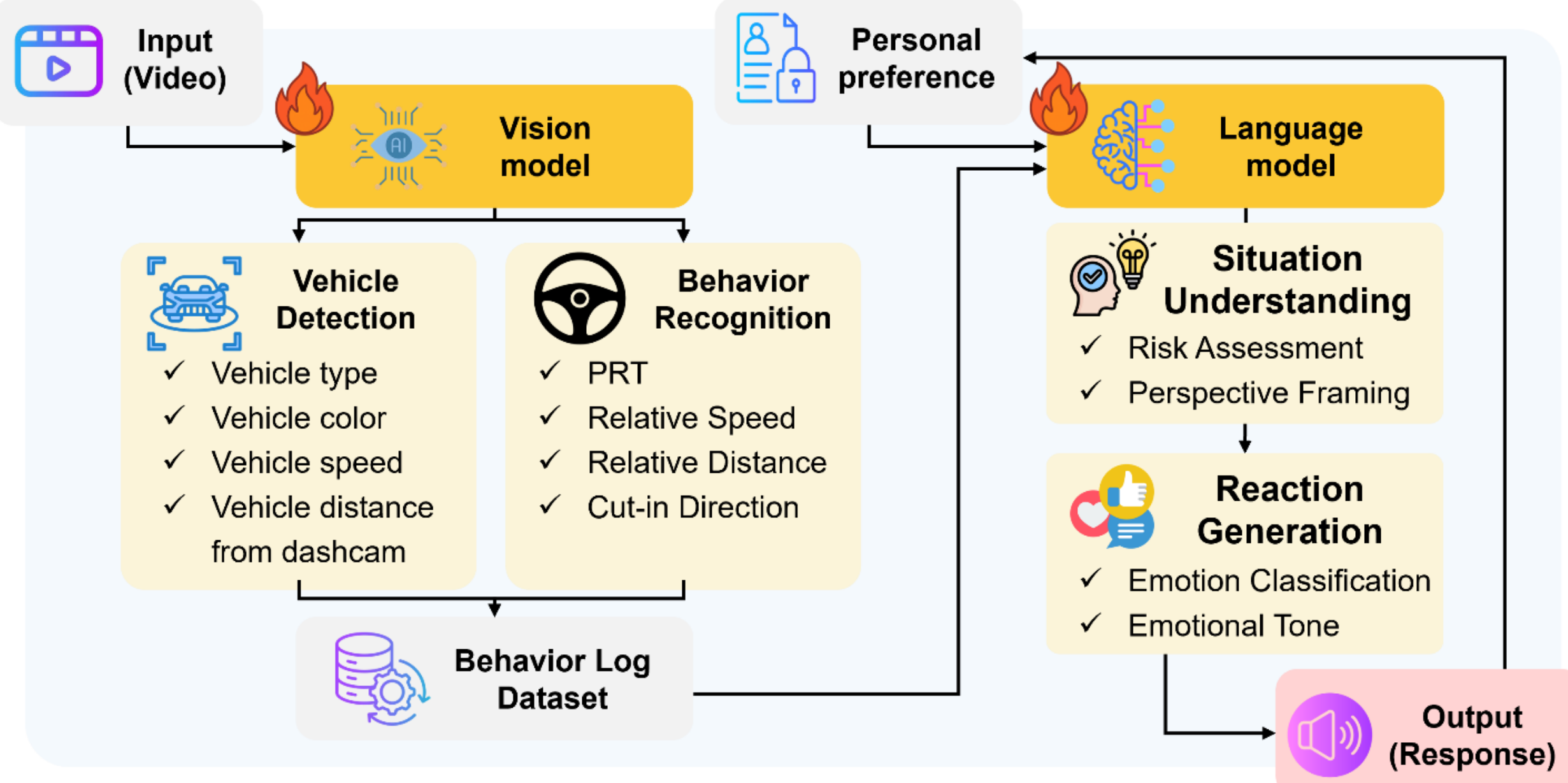

**Figure 1. Overview of the KYA Pipeline**

**Vision Model**

The vision module serves as the first stage in the KYA pipeline, transforming raw dashcam footage into structured spatiotemporal representations that can be processed by the downstream language model. It consists of two core components: vehicle detection and driving behavior recognition.

To enable reliable detection and tracking, a variety of computer vision models can be employed, including SSD, Faster R-CNN, and the You Only Look Once (YOLO) family of real-time object detectors (*30,31*). Among these, YOLOv8 is selected for its strong balance between detection accuracy and inference speed, making it particularly suitable for dynamic and unconstrained driving scenarios.

In this study, we adopt YOLOv8, pretrained on large-scale autonomous driving datasets such as BDD100K and nuScenes. We utilize all four YOLOv8 variants, i.e., nano (n), small (s), medium (m), and large (l), to explore trade-offs between model size and accuracy. To further optimize performance for dashcam-based applications, we fine-tune the YOLOv8n and YOLOv8s variants by adjusting several key hyperparameters. In particular, the confidence threshold is lowered from 0.5 to 0.3 to enhance the detection of small or partially occluded vehicles. For the medium and large variants (YOLOv8m and YOLOv8l), default settings are retained to preserve their baseline accuracy.

To ensure consistent input quality, video sampling rates are configured between 5 and 10 frames per second, depending on scene complexity. This setting maintains temporal continuity while reducing redundant frame processing and computational overhead.

*Vehicle detection*

Vehicle detection constitutes a foundational step in the KYA pipeline. In this stage, each vehicle appearing in the dashcam footage is detected and assigned a unique identifier. For every detected instance, the system extracts and records key attributes that capture both spatial and contextual information. These include the vehicle's image-plane coordinates (x, y), its estimated speed in kilometers per hour, and its distance from the ego vehicle, such as the driver's own vehicle equipped with the dashcam. The model also classifies the vehicle type (e.g., car, bus, or truck) and estimates its color. In addition, the environmental context is annotated, including weather conditions (such as clear or rainy) and the road type (e.g., urban street or highway) where the interaction occurs.

A major challenge in analyzing dashcam video lies in the variability introduced by camera angles, mounting positions, and lens distortion (*32*). To address these factors and ensure consistency in spatial measurements, we apply a normalization procedure. This reference functions as a scale anchor that preserves relative proximity relationships within each video. All other distance-related features are rescaled proportionally based on this reference. A similar normalization is applied to estimated velocities and spatial coordinates to maintain consistent relative motion patterns across varying camera perspectives. This design prioritizes scale-invariant proximity compression cues over absolute depth accuracy. All extracted features are compiled into a structured behavior log that represents the dynamic traffic scene on a frame-by-frame basis. This log serves as the foundational input for downstream analysis and emotion inference modules within the KYA system.

*Driving Behavior Recognition*

Driving behavior recognition in our system focuses on identifying vehicles that may pose a risk to ego vehicles, especially those making sudden, risky maneuvers such as cut ins. Rather than tracking all surrounding vehicles continuously, we streamline the process by analyzing key motion features like relative distance and speed. These features are directly derived from dashcam footage to efficiently detect potentially hazardous behaviors.

Since our dataset exclusively consists of cut-in scenarios, we assume that each scene contains at least one vehicle performing a risky maneuver toward the ego vehicle. For every frame, we calculate how rapidly and how closely each surrounding vehicle is approaching the ego vehicle. By analyzing these values over time, we identify the vehicle with the highest combination of proximity and closing speed. This vehicle is labeled as the aggressor and becomes the focal point for behavior analysis. In this setup, we are not defining cut-in events ourselves; rather, we focus on pinpointing the primary source of risk within scenes where a cut-in has already occurred.

Once the aggressor is identified, we extract its spatial and contextual attributes, including image position (x, y), estimated speed (kph), relative distance (meters), vehicle type, and color, based on the outputs of the detection module. In addition, we compute a risk indicator named PRT, which estimates the time it would take for the ego vehicle to reach the current position of the aggressor, assuming a constant speed. Unlike traditional surrogate safety metrics such as Time-to-Collision (TTC) or Post-Encroachment Time (PET), these measures rely on predicting the future motion of two interacting vehicles. TTC estimates the remaining time before a potential collision, and PET measures the time gap between two vehicles passing the same conflict point. In contrast, PRT is calculated solely from the ego vehicle's trajectory, making it suitable for single-camera dashcam environments where the future motion of surrounding vehicles is unknown. PRT reflects surrounding vehicle behavior through changes in relative distance and is highly sensitive to sudden proximity shifts. For example, when a vehicle abruptly cuts in front of the ego vehicle, the relative distance decreases sharply, resulting in a lower PRT value that signals an emerging proximity hazard. Its ego-centric formulation also allows PRT to be applied consistently in multi-vehicle situations. It offers a conservative yet informative measure of temporal proximity, particularly effective for identifying abrupt cut-in behaviors. The PRT is defined as:

$$PRT = d_{rel}/\left(v_{ego} + \varepsilon\right) \quad (1)$$

where $d_{rel}$ is the relative distance between the ego vehicle and the aggressor (in meters), and $v_{ego}$ is the current speed of the ego vehicle (kph). A small constant $\varepsilon$ is added to the denominator to prevent division by zero. Lower PRT values indicate that the ego vehicle is projected to reach the aggressor's current position within a short time, signaling elevated collision risk.

The extracted indicators are intended as relative behavioral risk cues derived from visual dynamics, rather than physically calibrated collision metrics. Specifically, this simple yet effective metric enables the system to quantify risky proximity patterns without relying on trajectory-level prediction or precise sensor calibration.

**Language Model**

The language model constitutes the second stage of the KYA pipeline, transforming structured behavioral data into emotionally expressive natural language responses. Alongside the behavior log, the module also receives user-specific emotional preference settings, which are collected through a preliminary survey. In this survey, participants selected their preferred response persona from a set of emotional tones, including neutral, mildly annoyed, moderately annoyed, highly angry, humorous, generous, and analytical. These personas guide the linguistic and emotional style of the system's verbal reactions, allowing responses to be tailored to the driver's personality and mood. Once the vision module has identified the aggressor and extracted its behavioral attributes, this information, along with the user's persona selection, is passed to the language module for high-level reasoning and feedback generation. We utilize multiple state-of-the-art LLMs, including ChatGPT-4o (*33*), Claude 3 (*34*), Gemini 2.5 (*35*), and Copilot (*36*), to support this module.

The language module performs two main functions: situation understanding and reaction generation. In the first stage, the model interprets the traffic scenario by analyzing the structured behavior log, which includes variables such as the aggressor's relative distance, relative speed, PRT, vehicle type, color, and the cut-in direction (e.g., left or right). These inputs provide critical cues for assessing the severity of the interaction and the level of risk it poses to the ego vehicle. By evaluating how abruptly the vehicle approached, from which side, and how quickly the time-to-impact is decreasing, the model builds a contextual understanding of the event. This mirrors how human drivers perceive and emotionally react to nearby threats.

In the second stage, the model generates a natural language response that aligns with both the situation and the user's selected emotional tone. The final response is then synthesized into speech using a text-to-speech (TTS) system, which adapts vocal characteristics to match the chosen persona, for example, using a calm tone for a neutral setting or exaggerated inflection for a humorous one.

This two-stage LLM architecture allows the KYA system to translate sensor-level observations into emotionally rich, context-aware feedback. By integrating perception with personality, it creates a driving assistant that not only understands the road but also speaks in a voice that feels familiar, expressive, and human.

**DATA**

We used two distinct data sources to evaluate the proposed framework. First, for the vision module, we curated real-world dashcam videos from YouTube (https://www.youtube.com). Using search terms related to sudden cut-ins, we selected 20 videos, including both standard and short-form content that depict high-risk driving scenarios. These videos were processed to extract frame-by-frame vehicle dynamics, resulting in a total of 5,019 frames across 20 ego-vehicle sequences. Among them, 1,189 frames were identified as high-risk frames associated with aggressive cut-in behavior. The selected videos represent independent real-world interaction contexts with diverse traffic and behavioral characteristics, providing sufficient variability to evaluate the proposed framework. Although the dataset comprises 20 video sequences, each sequence reflects naturally occurring driving interactions with temporally continuous behavioral dynamics. These event-level samples provide diverse interaction contexts sufficient for examining how perceptual cues are translated into structured behavioral indicators and subsequently into language responses.

Second, to capture user preferences for language-based emotional responses, we conducted a survey with 108 participants (53 male and 55 female). The survey consisted of two parts: a preference profiling stage and a model evaluation stage. In the profiling section, participants reported their gender, age, and preferred emotional tone, selecting from seven predefined persona types: Neutral, Light Anger,

Noticeable Anger, Intense Anger, Analytical, Humorous, and Magnanimous. For the model evaluation, we assessed the emotional alignment performance of four LLMs: ChatGPT-4o, Claude 3, Gemini 2.5, and Copilot. Each model was evaluated under both zero-shot and few-shot prompting conditions. The evaluation consisted of three components:

- Persona-wise preference: For each persona, participants were shown four anonymized responses (one per model) and asked to select the most appropriate one based on emotional tone and stylistic fit.
- Sentence-level model choice: Participants selected the best response from four options within each persona category.
- Model-wise rating: Participants reviewed a full set of responses from a given model and rated it on a 5-point Likert scale based on fluency, emotional alignment, and persona consistency.

In summary, the video data served as input to the vision module, which produced structured outputs such as relative speed, distance, and PRT. These outputs, along with survey-derived user profiles and preferences, were used as input to the language module. This dual-input design supports an efficient pipeline where real-world visual cues and human preference signals are integrated to produce contextually and emotionally appropriate responses.

# RESULTS

## Results of Vision Model

### *Performance of Vehicle Detection*

To evaluate vehicle detection and tracking performance, we applied the YOLOv8 object detector to real-world dashcam footage collected from public video platforms, focusing on high-risk driving scenes such as cut-ins and close merges. We tested four different YOLOv8 variants (YOLOv8n, YOLOv8s, YOLOv8m, and YOLOv8l) to analyze the trade-off between detection accuracy and inference efficiency.

**Table 1** summarizes the quantitative performance of each model based on widely used object detection and tracking metrics (*21-24*), including precision, recall, F1 score, mean Average Precision at IoU threshold 0.5 (mAP@0.5), ID switches, Multi-Object Tracking Accuracy (MOTA), and Intersection over Union (IoU). These metrics are among the most widely used for evaluating object detection and tracking performance.

- Precision: the proportion of detected vehicles that are correctly identified (low false positives).
- Recall: the proportion of actual vehicles that are successfully detected (low false negatives).
- F1 Score: the harmonic mean of precision and recall, balancing both detection accuracy and completeness.
- mAP@0.5: the mean Average Precision at an IoU threshold of 0.5, measuring overall detection accuracy across all objects.
- ID Switches: the number of times a detected object is incorrectly re-identified across frames.
- MOTA (Multi-Object Tracking Accuracy): an aggregated metric that combines false positives, false negatives, and ID switches to reflect overall tracking performance.
- IoU (Intersection over Union): the degree of overlap between predicted and ground-truth bounding boxes, indicating spatial localization accuracy.

Among the evaluated models, YOLOv8l achieved the best overall performance. It recorded the highest recall (0.869), F1 score (0.921), and mAP@0.5 (0.998), along with the lowest number of ID switches (1), indicating strong temporal consistency in multi-object tracking. In terms of bounding box alignment, it also showed the highest average IoU score (0.988), suggesting robust spatial precision. Lighter models, i.e., YOLOv8n and YOLOv8s demonstrated greater computational efficiency but exhibited slightly lower recall values (0.737 and 0.762, respectively) and reduced tracking stability (MOTA: 0.715 and 0.745). Nevertheless, their high precision (0.979 and 0.9822) and low ID switch counts (3 and 2, respectively) suggest that even smaller models can deliver acceptable performance for real-time deployment.

For detecting vehicles that pose immediate threats, such as sudden cut-ins in proximity, YOLOv8s and larger variants consistently capture critical agents with high reliability. YOLOv8s is sufficient for emotionally responsive applications focused on near-field risk perception. In contrast, when broader detection coverage is required across all visible traffic participants, as in full-scene analysis or large-scale tracking, YOLOv8m offers a favorable balance between detection fidelity and computational cost. These results confirm the viability of YOLOv8-based detection and tracking as a foundational module for high-level scene understanding. The choice of model should depend on the specific application requirements, particularly the trade-off between real-time constraints and accuracy. The performance of vehicle detection models is summarized in **Table 1**.

**Table 1 Performance of vehicle detection**

| Model | YOLOv8n | YOLOv8s | YOLOv8m | YOLOv8l |
|---|---|---|---|---|
| Precision | 0.979 | 0.982 | 0.994 | 0.977 |
| Recall | 0.737 | 0.762 | 0.771 | 0.869 |
| F1 Score | 0.841 | 0.858 | 0.868 | 0.921 |
| mAP@0.5 | 0.985 | 0.991 | 0.989 | 0.998 |
| ID Switches | 3 | 1 | 1 | 1 |
| MOTA | 0.715 | 0.745 | 0.768 | 0.854 |
| Avg. IoU | 0.86 | 0.969 | 0.986 | 0.988 |

**Figure 2** illustrates the qualitative performance of YOLOv8s in detecting vehicles across a range of real-world driving scenarios, including varying road environments and lighting conditions. The selected examples encompass four representative contexts: sunny daytime, nighttime, urban roads, and freeways. In daytime conditions, YOLOv8s accurately detects vehicles even under strong sunlight, glare, and shadow, and successfully captures critical events such as sudden cut-ins. At night, the model continues to perform reliably despite reduced visibility, using visual cues like headlights and taillights to detect nearby and distant vehicles, including those partially occluded. In complex urban scenes with dense traffic, diverse vehicle types, and frequent occlusions, YOLOv8s demonstrates strong detection performance for both moving and stationary vehicles. On freeways, it consistently identifies vehicles at high speeds and across lanes, handling motion blur and scale variation effectively.

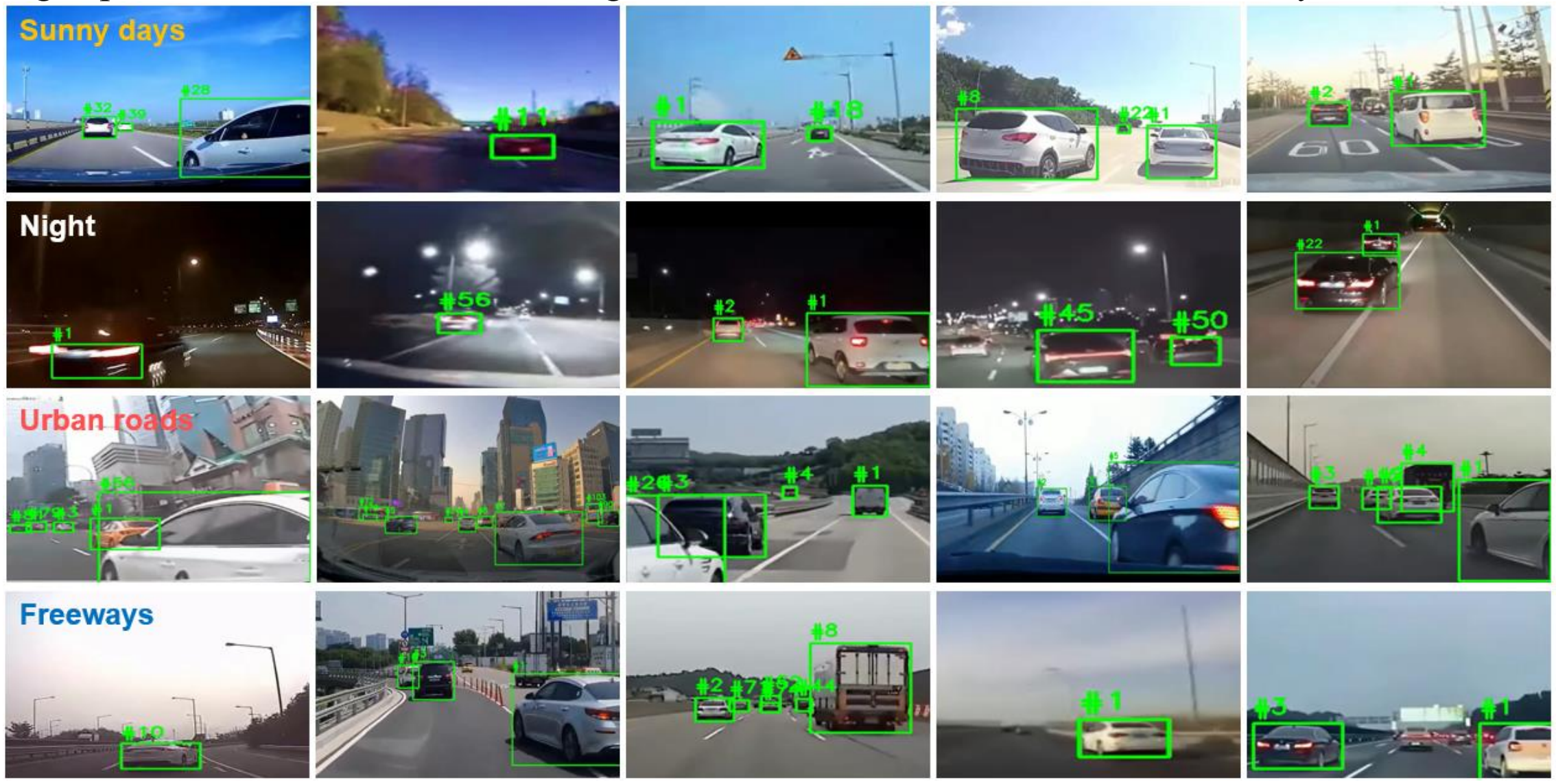

**Figure 2.** Examples of vehicle detection using YOLOv8s across various traffic scenes, including sunny days, night, urban roads, and freeways.

*Driving Behavior Estimation*

To identify instances of risky driving, we extracted driving behavior metrics from dashcam-tracked vehicle data. For each frame, the system computed relative distance, relative speed, and PRT between the ego vehicle and surrounding vehicles. We focused on identifying the single most threatening vehicle per frame by selecting the one with the shortest PRT value, which captures the urgency of potential collision based on both distance and speed. This inverse approach enabled us to isolate frames associated with heightened risk from the ego vehicle's perspective.

Out of the 20 vehicles observed, a total of 1,189 frames were identified as containing risky interactions. These frames were selected based on thresholding relative behavior indicators and verified through visual inspection to ensure contextual relevance (e.g., sudden cut-ins or close merges).

**Table 2** presents descriptive statistics for the three key behavioral indicators measured across these frames. The average relative distance between the ego vehicle and the closest threat was 5.43 meters, with a minimum observed value of just 0.5 meters. Relative speed averaged 15.55 km/h but reached a maximum of 168.18 km/h in extreme cases, indicating high-speed incursions. These extreme values reflect instantaneous relative velocity estimates derived from frame-level positional changes rather than calibrated sensor measurements.

The mean Projected Reach Time was 1.80 seconds, with over 75% of observed threats occurring within a 3-second threshold. According to the National Highway Traffic Safety Administration's "three-second rule," this duration is considered the minimum safe headway for following distance (*37*). However, given the relatively low mean and high variability of PRT (standard deviation: 1.35 seconds), the actual time available for evasive response is often much shorter than the guideline suggests. This highlights that, in many cases, drivers are confronted with dangerously narrow reaction windows, especially in scenarios involving rapid cut-ins by high-speed vehicles.

Extreme relative speed values (e.g., up to 168 km/h in Table 2) were observed in several cut-in events. These values reflect instantaneous relative velocity estimates derived from frame-level positional changes rather than calibrated vehicle speed measurements. Because the system operates on visually inferred dynamics without access to sensor-based ground truth, these values represent interaction intensity rather than physically validated speed. Manual inspection of the corresponding video segments confirmed that these cases corresponded to rapid lane intrusions with sharp proximity changes, rather than object detection errors. Therefore, the extreme values should be interpreted as high-magnitude behavioral interaction signals rather than absolute speed measurements.

Our results show that many threatening vehicles violated this rule, particularly during abrupt lane changes or risky merging maneuvers. These findings demonstrate the model's effectiveness in capturing risky driving behaviors based on spatiotemporal proximity and relative motion patterns.

**Table 2 Descriptive Statistics of Risky Driving Interactions**

| Metric | Mean | Std | Min | 25% | 50% | 75% | Max |
|---|---|---|---|---|---|---|---|
| Relative distance (m) | 5.43 | 2.80 | 0.50 | 4.15 | 5.30 | 6.55 | 12.80 |
| Relative speed (km/h) | 15.55 | 16.96 | 0.00 | 5.40 | 10.80 | 21.60 | 168.18 |
| Projected Reach Time (s) | 1.80 | 1.35 | 0.02 | 0.67 | 1.44 | 2.80 | 6.03 |

**Figure 3** illustrates the empirical distribution of key behavioral indicators observed during risky driving interactions: relative distance, relative speed, and PRT. These distributions offer deeper insight into the dynamics of risky maneuvers beyond aggregate summary statistics. In **Figure 3(a)**, the histogram of relative distance shows a concentration of interactions within the 3 to 6-meter range, with a peak around 4.5 meters. This suggests that most cut-ins occur at moderately close distances. However, the long tail extending beyond 10 meters indicates that some vehicles initiated lane changes from farther away, likely at higher speeds. In close cut-in events, the aggressor often approaches the ego vehicle to within approximately 0.5 meter, which explains the elevated frequency observed near this distance. In **Figure 3(b)**, the distribution of relative speed reveals that most interactions occurred at speed differentials below 40 km/h. Nevertheless, the presence of a long tail reaching up to 130 km/h highlights rare but severe instances of high-speed aggression. Although infrequent, these extreme cases pose serious safety risks due to the minimal time available for evasive response. In **Figure 3(c)**, the PRT distribution exhibits a pronounced right skew, with most interactions falling below the 3-second safety threshold. A substantial proportion of events are clustered between 0.5 and 2 seconds, indicating that drivers were often confronted with extremely limited reaction time. This observation reinforces the conclusion that many risky interactions violate the U.S. Department of Transportation's recommended “three-second rule.”

Together, these distributional patterns reveal that risky driving behavior is not only frequent but also diverse in its intensity and timing. They underscore the importance of accounting for both typical and rare high-risk scenarios in downstream behavioral modeling and emotional response generation.

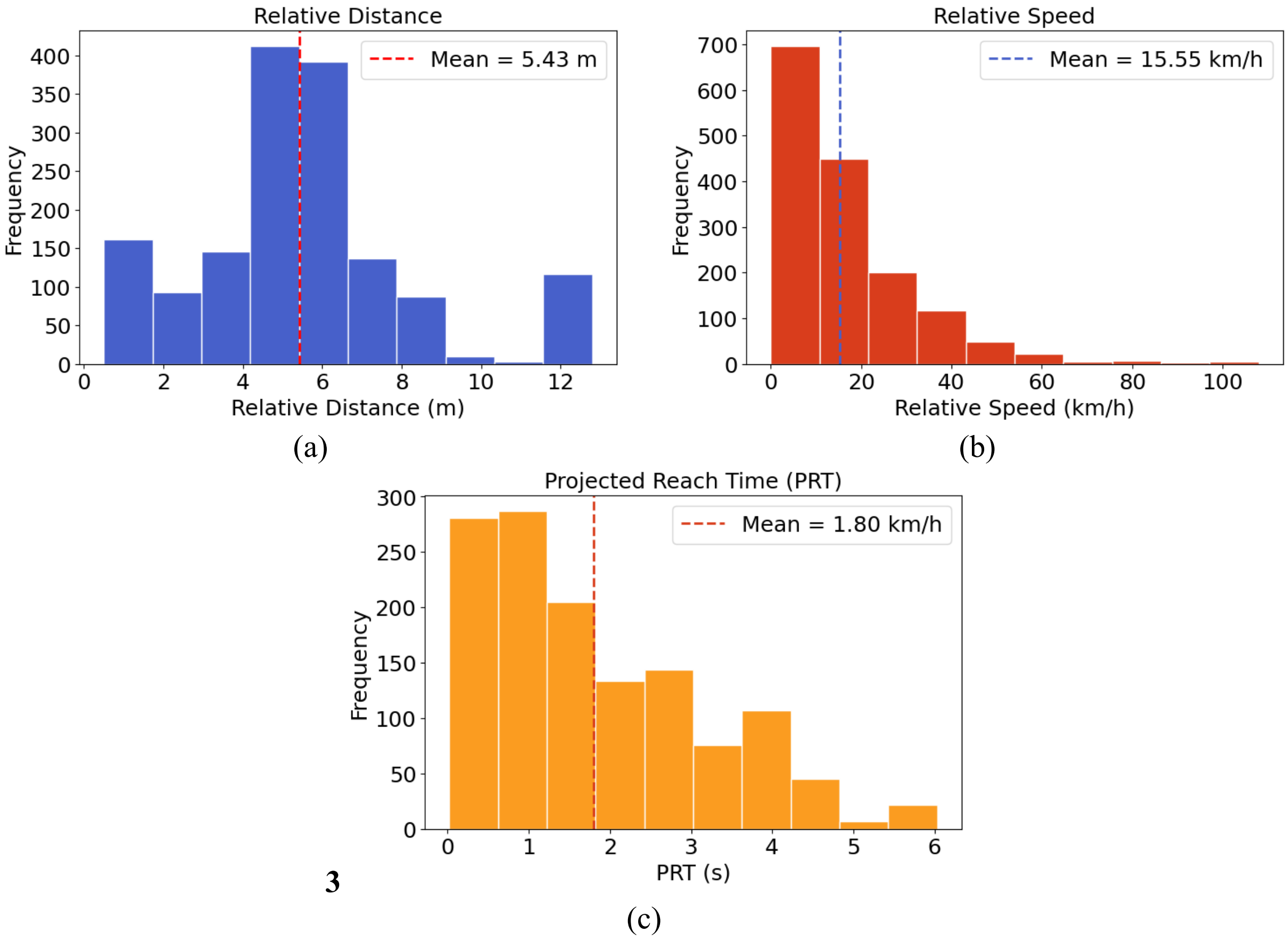

**Figure 3. Distribution of key behavioral risk indicators: (a) relative distance from the ego vehicle (m), (b) relative speed (km/h), and (c) PRT (seconds).**

## Results of Language Model

*Performance of Emotional Response Generation*

We evaluated the emotional alignment performance of four vision-language models: ChatGPT-4o, Claude 3, Gemini 2.5, and Copilot. The evaluation focused on their ability to generate situationally appropriate responses across seven predefined persona types: Neutral, Light Anger, Noticeable Anger, Intense Anger, Analytical, Humorous, and Magnanimous.

**Table 3** summarizes the percentage of participants who selected each model's response as the top choice across the seven persona types. These selection rates indicate how frequently a model's output was perceived as the most appropriate for each emotional style. The lower section of the table reports two aggregate metrics: average preference and overall score. The average preference represents the simple mean of each model's selection rates across all personas.

The overall score integrates three components to assess model performance: (1) the proportion of times a model's response was selected for each persona type, (2) the relative importance of each persona derived from observed user preferences (e.g., neutral: 0.083, light anger: 0.083, noticeable anger: 0.133, intense anger: 0.110, analytical: 0.232, humorous: 0.265, magnanimous: 0.094), and (3) the overall average rating of the KYA system under zero-shot and few-shot conditions.

Specifically, model-level selection proportions were weighted by persona importance to compute a weighted preference score for each model. Because only the overall mean rating was available for each prompting condition (3.35 for zero-shot and 3.49 for few-shot), we proportionally distributed this observed average across models according to their weighted preference scores. This approach preserves the observed overall rating while allocating it in a manner consistent with the underlying preference structure.

The results showed that ChatGPT-4o consistently outperformed other models in both zero-shot and few-shot conditions. It achieved the highest selection rates in analytical (60.2%) and neutral (47.8%) personas under zero-shot, and continued to lead in humorous (42.2%) and analytical (41.6%) personas under few-shot. Claude 3 showed competitive performance in emotionally charged styles, particularly light anger (36.8%) and intense anger (36.4%) in zero-shot. Copilot also exhibited relative strengths in neutral and intense anger personas, especially under few-shot conditions.

In terms of average preference, Claude 3 ranked highest in the zero-shot setting, followed by ChatGPT-4o, Copilot, and Gemini. In the few-shot setting, ChatGPT-4o led, followed by Claude 3, Gemini, and Copilot. For overall integrated scores, ChatGPT-4o obtained the highest estimated values in both prompting modes (4.10 for zero-shot and 4.29 for few-shot), followed by Claude 3 (3.85 and 3.98, respectively). Gemini and Copilot recorded lower integrated scores.

These findings indicate that all models are generally capable of generating emotionally aligned responses, as shown by their similar average preference scores. However, when factoring in persona-specific preferences and qualitative ratings, clear differences emerge.

To further examine whether the observed differences in model selection are statistically meaningful, a chi-square test was conducted on the raw selection frequencies. The results indicate a statistically significant difference across models ($\chi^2 = 12.20$, $p = 0.0067$). However, since participants were allowed to select multiple responses, the independence assumption is partially relaxed. Accordingly, the results should be interpreted as supportive rather than definitive statistical evidence.

**Table 3 Performance of VLMs**

| | Zero shot (Avg. 3.35 score) | | | | Few shot (Avg. 3.49 score) | | | |
|---|---|---|---|---|---|---|---|---|
| | ChatGPT-4o | Claude | Gemini | Copilot | ChatGPT-4o | Claude | Gemini | Copilot |
| Neutral (%) | 47.8 | 45.6 | 2.2 | 4.4 | 24.2 | 21.8 | 21.8 | 32.2 |
| Light anger (%) | 11.5 | 36.8 | 11.5 | 40.2 | 34.1 | 38.6 | 19.3 | 8.0 |
| Noticeable anger (%) | 14.9 | 19.5 | 33.4 | 32.2 | 19.1 | 27.0 | 29.2 | 24.7 |
| Intense anger (%) | 28.4 | 36.4 | 12.5 | 22.7 | 23.6 | 27.0 | 18.0 | 31.4 |
| Analytical (%) | 60.2 | 20.5 | 8.0 | 11.3 | 41.6 | 15.7 | 18.0 | 24.7 |
| Humorous (%) | 27.0 | 38.2 | 24.7 | 10.1 | 42.2 | 18.9 | 21.1 | 17.8 |
| Magnanimous (%) | 19.5 | 21.8 | 35.6 | 23.1 | 23.9 | 25.0 | 31.8 | 19.3 |
| Avg. Preference (%) | 29.9 | 31.3 | 18.3 | 20.6 | 29.8 | 24.9 | 22.7 | 22.6 |
| Overall Score (1 to 5) | 4.10 | 3.85 | 2.76 | 2.69 | 4.29 | 3.98 | 2.88 | 2.81 |

*User Preferences and Personalization Insights*

To evaluate user preferences for emotionally expressive in-vehicle assistants, we conducted a supplementary survey-based analysis. **Figure 4** presents the summarized findings across four key dimensions: persona style, voice type, system adoption intention, and language model preference. The responses were blindly evaluated – participants were unaware of which LLM produced each.

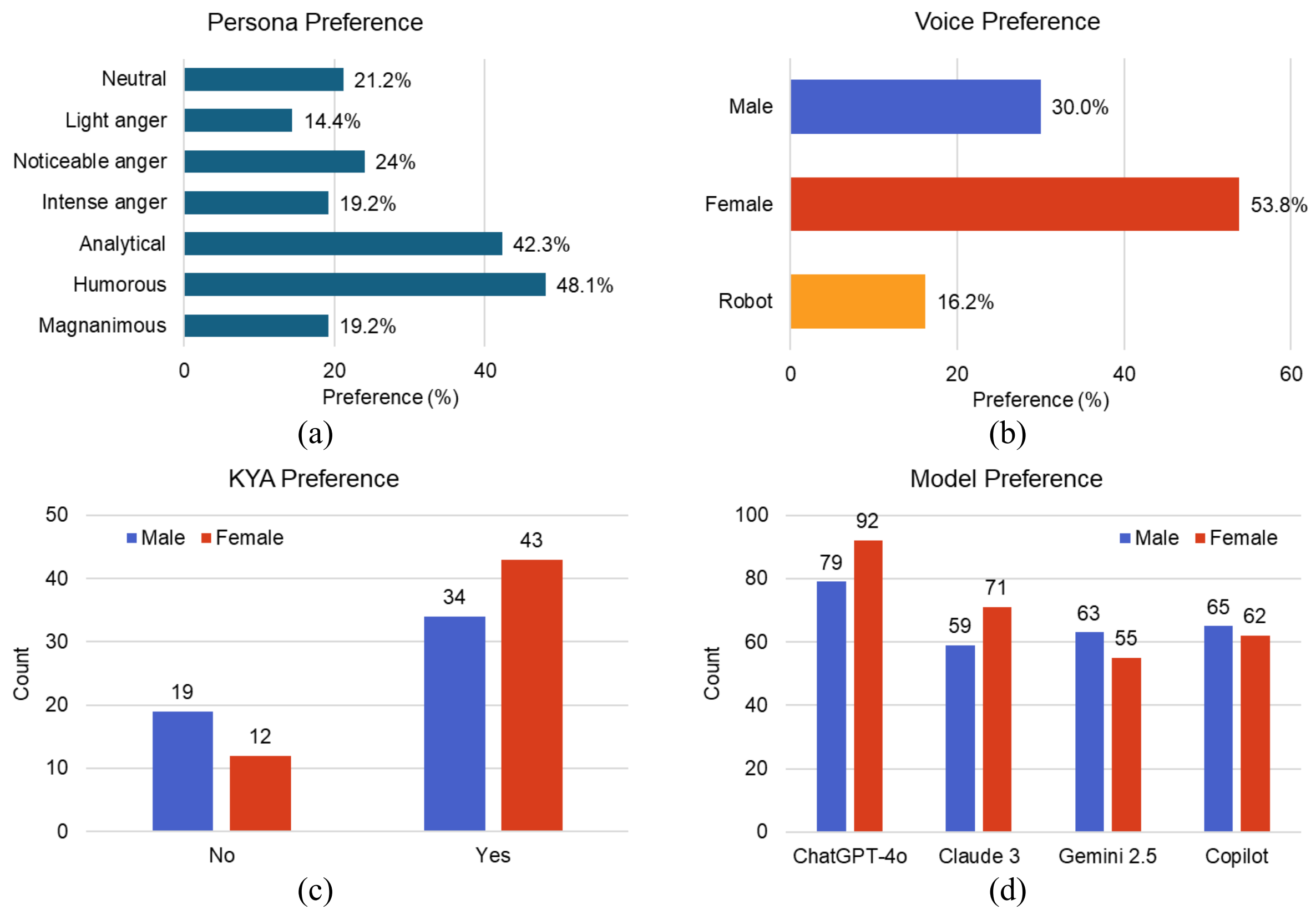

**Figure 4. Survey results on user preferences: (a) persona style, (b) voice type, (c) KYA adoption, (d) language models**

First, we assessed participants' preferences for various assistant personas, which were used as weighting factors in the overall score computation described earlier. As shown in **Figure 4(a)**, humorous (48.1%) and analytical (42.3%) personas were most favored, indicating a clear preference for responses that balance rationality with emotional levity. In contrast, personas exhibiting noticeable (24.0%) or intense anger (19.2%) received lower ratings, suggesting users generally avoid interactions that mimic risky or overly emotional behavior. This finding reinforces the importance of emotional alignment over emotional intensity in human–AI interaction design.

Second, voice preference results, as shown in **Figure 4(b)**, revealed a strong inclination toward female voices (53.8%), followed by male (30.0%) and robotic voices (16.2%). The preference for female voices may be attributed to perceived qualities such as warmth, empathy, or reassurance—traits that are particularly valued in emotionally sensitive or stressful driving scenarios. Interestingly, robotic voices were not well received, possibly due to their perceived lack of authenticity or emotional nuance, which could hinder trust and engagement in real-time driving contexts. This highlights the importance of incorporating anthropomorphic design elements in voice-based agents such as KYA.

Third, we measured participants' willingness to adopt the KYA system, as shown in **Figure 4(c)**. A substantial majority expressed interest, with 43 female and 34 male respondents indicating they would use such a system. This gender difference suggests that female users may be more open to emotionally expressive AI assistants, potentially due to greater receptiveness to emotionally intelligent support during high stress driving conditions. The positive response across both genders, however, signals a general user readiness for emotionally aligned in-vehicle AI companions.

Lastly, we explored model preference in a blind setting, where participants evaluated system outputs without knowing the underlying model, as shown in **Figure 4(d)**. ChatGPT-4o received the highest vote count from both male and female users, confirming its superior perceived quality and emotional alignment. Interestingly, gender-specific differences emerged in secondary preferences: female users favored Claude 3, followed by Copilot and Gemini 2.5, while male users preferred Copilot first, then Gemini 2.5, and Claude 3 last. These patterns suggest that female users may value emotionally articulate, nuanced responses (as associated with Claude 3), whereas male users may prioritize consistency or pragmatic styles. Additionally, both male and female participants showed a general preference for female voices, and female participants tended to favor humorous response styles more than other groups.

These findings highlight the potential importance of model-level personalization in multi-agent systems designed for diverse user populations. Rather than pointing to a single fixed emotional tone, the results indicate that drivers rely on different styles depending on traffic conditions, stress levels, and individual mood. While a representative scenario helps reveal overall preference patterns, our analysis suggests that effective deployment may benefit from emotional tones that adapt dynamically to each user.

*Generated Emotional Reactions*

To evaluate the perceived quality of emotional responses, we conducted a sentence-level preference test for each of the seven persona styles. Participants were presented with four anonymized responses. Each response was generated by ChatGPT-4o, Claude 3, Gemini 2.5, or Copilot. Since each model generates emotional reactions in distinct ways, the comparison was conducted with two objectives. First, it assesses users' overall preferences for different forms of emotional expression. Second, it establishes a basis for determining which model is most suitable for real-world deployment in in-vehicle systems. They were asked to select the one they found most appealing within each stylistic context.

**Figure 5** presents the results of the sentence-level preference test, which evaluated model performance across seven emotional personas. ChatGPT-4o showed strong performance overall, particularly in humorous and analytical styles. For the humorous persona, a response using metaphorical language and informal tone such as "Whoa, guess he thought this was GTA and lanes were optional. Pure comedy!" was preferred by 42.2% of participants. In the analytical style, a structured and data-driven message like "A car suddenly changed lanes from the left without signaling, with a speed difference of over 25 km/h. We'll slow down and maintain a 50-meter safe distance" received 41.6% of the votes, demonstrating the model's ability to generate contextually appropriate and persona-consistent output.

In emotionally intense scenarios, Claude 3 performed competitively. One of its intense anger responses, "Hey, you XXX! Are you that eager to cause a crash?!", was selected by 27.0% of participants for effectively capturing the intended emotional tone. In this context, it outperformed ChatGPT-4o. Claude also showed notable performance in the noticeable anger and humorous categories, indicating its strength in handling expressive personas.

While ChatGPT-4o and Claude 3 achieved the highest overall scores, Gemini 2.5 and Copilot also demonstrated strengths in specific styles. Gemini 2.5 was well-received in the noticeable anger category with its assertive response, "Hey! What's with the sudden cut-in? Still, let's calm down. Staying safe is the priority," which participants found emotionally appropriate and balanced. It also performed well in the magnanimous persona with the calm and empathetic message, "Haha, someone's clearly in a hurry. Drive safe out there," suggesting its potential for generating considerate reactions.

Copilot showed strengths in both the neutral and intense anger styles. In the neutral category, its response, "That car cut in a bit suddenly. But let's not panic and keep driving steadily," conveyed a composed yet mildly annoyed tone reflective of real-world driving. In the intense anger setting, its line, "XXX. You think you own the road?! Fine, race your way to the afterlife!" illustrated its ability to combine emotional intensity with evaluative commentary. These results suggest that even less-preferred models can generate compelling outputs when well-aligned with specific emotional personas.

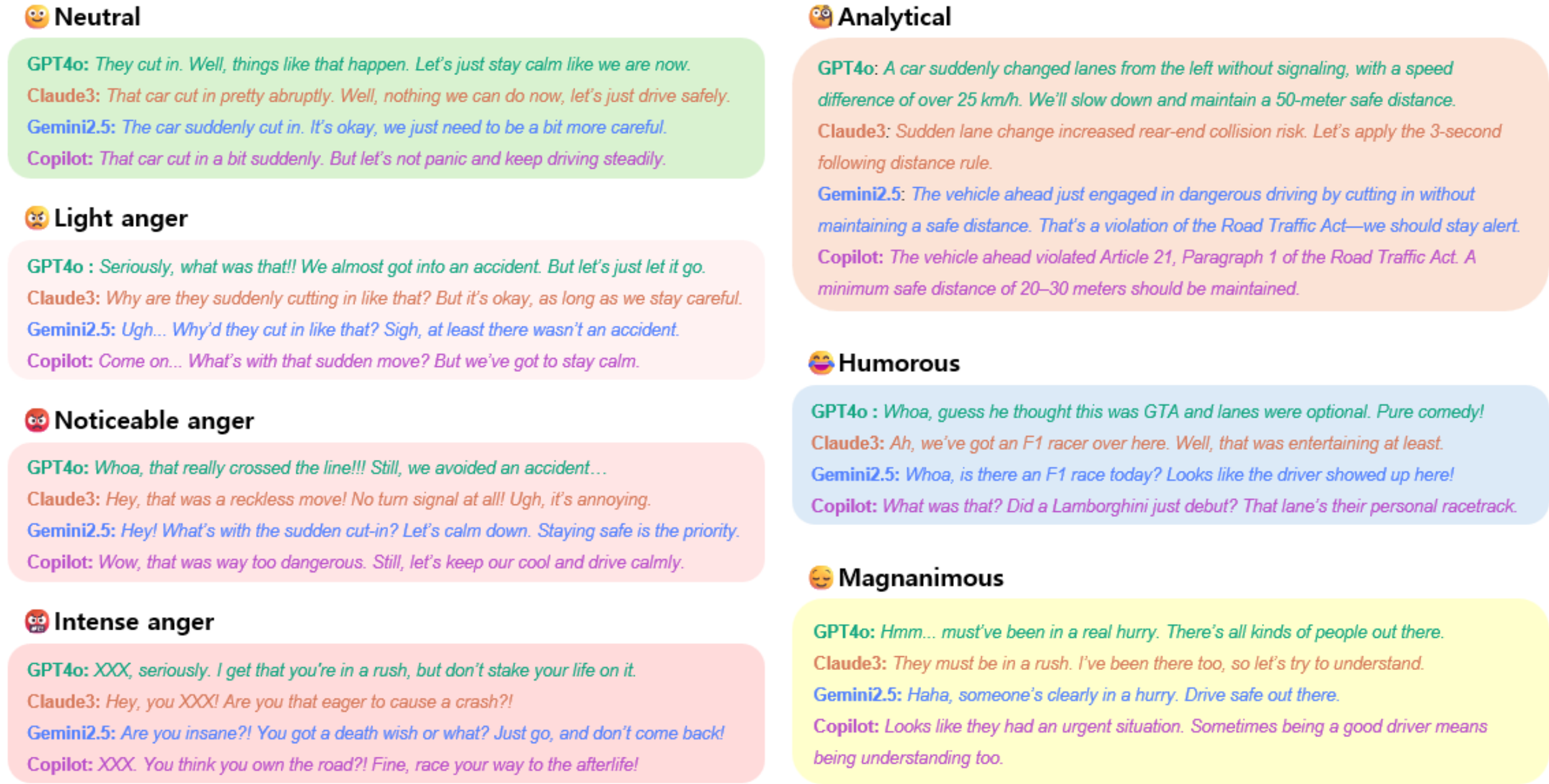

**Figure 5. Responses generated by each model across seven emotional personas.**

## CONCLUSIONS

This study proposed KYA, a modular vision-language framework designed to detect risky driving behaviors, particularly sudden cut-ins, and generate emotionally expressive, persona-aligned verbal reactions. The system bridges the perceptual capabilities of real-time object detection with the reasoning power of large language models, enabling a novel form of affect-aware driver support. By emulating how human drivers emotionally respond to risky interactions, KYA contributes to a growing body of work on humanized AI in the context of intelligent transportation systems.

The primary contribution of this work is the establishment of an integrated perception–behavior–emotion framework that bridges driving risk abstraction and persona-aligned language generation, enabling affect-aware driver interaction. Specifically, our contributions are threefold. First, we developed an efficient and interpretable vision-language pipeline in which perception and reasoning components are clearly modularized. The vision and language modules each operate within their respective domains of strength, and the connection between them is facilitated through structured inputs derived from spatiotemporal driving cues such as relative distance, speed, and PRT. Unlike existing VLM-based systems focused on situational reasoning or task-specific guidance, KYA is among the first to produce emotionally expressive responses grounded in real-time driving scenarios.

Second, the vision module was carefully calibrated and tested using YOLOv8 models of varying complexity. Our evaluation demonstrated that mid-range models (e.g., YOLOv8s and YOLOv8m) strike an optimal balance between accuracy and processing speed, making them suitable for real-time applications. Moreover, we introduced a reproducible method for identifying risky cut-in events directly from unconstrained dashcam videos, using behavior indicators like PRT to quantify risk. Because these risk patterns are learned beforehand, the same criteria can be applied in real time to detect emerging hazardous behaviors. This design bridges the gap between low-level perception and high-level affective reasoning.

Third, we implemented a language model that adapts its responses based on user preferences. Through a structured survey of 108 participants, we identified emotional style preferences and evaluated the alignment performance of four large language models. The results show that emotionally tailored responses, especially in humorous or analytical styles, were consistently preferred. Additionally, this study is the first to directly integrate user-defined emotional personas into the response-generation process of an LLM-based assistant, enabling granular personalization.

Although the empirical evaluation was limited to sudden cut-in events, the framework is indicator-driven and behavior-agnostic in structure. Additional risky behaviors, such as tailgating or sudden braking, can be incorporated through the definition of corresponding behavioral metrics, such as time headway or longitudinal deceleration thresholds. Several limitations remain. All evaluations were conducted offline using pre-recorded dashcam footage and controlled user studies, and real-time in-vehicle validation has not yet been achieved. The participant sample was also relatively narrow in demographic scope. Future work will focus on real-world deployment, expanded behavioral categories, and broader user validation to examine longitudinal safety and usability impacts. In addition, the proposed framework is expected to be applicable to transportation planning and policy contexts by supporting data-driven safety assessment and intervention design (38-46).

## AUTHOR CONTRIBUTIONS

The authors confirm contribution to the paper as follows: study conception and design: H. Choi, E. H. Lee, and Z. Tu; data collection: H. Choi, E. H. Lee, and Z. Tu; analysis and interpretation of results: H. Choi, E. H. Lee, and Z. Tu; draft manuscript preparation: H. Choi, E. H. Lee, and Z. Tu. All authors reviewed the results and approved the final version of the manuscript.

## CONFLICT OF INTEREST

The authors declare that there is no conflict of interest regarding the publication of this paper.